\documentclass{article}

\usepackage{microtype}
\usepackage{graphicx}
\usepackage{subfigure}
\usepackage{booktabs} 

\usepackage{multicol}
\usepackage{multirow}
\usepackage{hyperref}

\usepackage{tikz}

\usepackage{colortbl}
\definecolor{Gray}{gray}{0.9}
\definecolor{LightCyan}{rgb}{0.75,1,1}
\definecolor{Yellow}{rgb}{1, 0.7, 0.2}

\usepackage[accepted]{icml2025}

\usepackage{amsmath}
\usepackage{amssymb}
\usepackage{mathtools}
\usepackage{amsthm}
\usepackage{float}
\usepackage{enumitem}
\usepackage[capitalize,noabbrev, nameinlink]{cleveref}

\Crefname{assumption}{Assumption}{Assumptions}

\theoremstyle{plain}
\newtheorem{theorem}{Theorem}[section]

\theoremstyle{definition}

\newtheorem{assumption}[theorem]{Assumption}
\theoremstyle{remark}

\usepackage{rotating}

\usepackage[textsize=tiny]{todonotes}

\icmltitlerunning{Compressed Chain of Thought}

\usepackage{dblfloatfix}    

\begin{document}

\setlength{\abovedisplayskip}{1em}
\setlength{\belowdisplayskip}{1em}

\twocolumn[
\icmltitle{Compressed Chain of Thought: Efficient Reasoning through \\Dense Representations}



\icmlsetsymbol{equal}{*}

\begin{icmlauthorlist}
\icmlauthor{Jeffrey Cheng}{jh}
\icmlauthor{Benjamin Van Durme}{jh}

\end{icmlauthorlist}

\icmlaffiliation{jh}{Department of Computer Science, Johns Hopkins University, Baltimore, US}

\icmlcorrespondingauthor{Jeffrey Cheng}{jcheng71@jh.edu}

\icmlkeywords{Machine Learning, ICML}

\vskip 0.3in
]



\printAffiliationsAndNotice{} 

\newcolumntype{A}{ >{\centering\arraybackslash} m{0.17\textwidth} }
\newcolumntype{B}{ >{\centering\arraybackslash} m{0.09\textwidth} }
\newcolumntype{C}{ >{\centering\arraybackslash} m{0.15\textwidth} }

\begin{abstract}

Chain-of-thought (CoT) decoding enables language models to improve reasoning performance at the cost of high generation latency in decoding. Recent proposals have explored variants of \textit{contemplation tokens}, a term we introduce that refers to special tokens used during inference to allow for extra computation. Prior work has considered fixed-length sequences drawn from a \textit{discrete} set of embeddings as contemplation tokens. Here we propose Compressed Chain-of-Thought (CCoT), a framework to generate \textit{contentful} and \emph{continuous} contemplation tokens of variable sequence length. The generated contemplation tokens are compressed representations of explicit reasoning chains, and our method can be applied to off-the-shelf decoder language models. Through experiments, we illustrate how CCoT enables additional reasoning over dense contentful representations to achieve corresponding improvements in accuracy. Moreover, the reasoning improvements can be adaptively modified on demand by controlling the number of contemplation tokens generated.
\end{abstract}

\section{Introduction}
\begin{figure*}[t]
    \centering
    \includegraphics[width=0.95\linewidth]{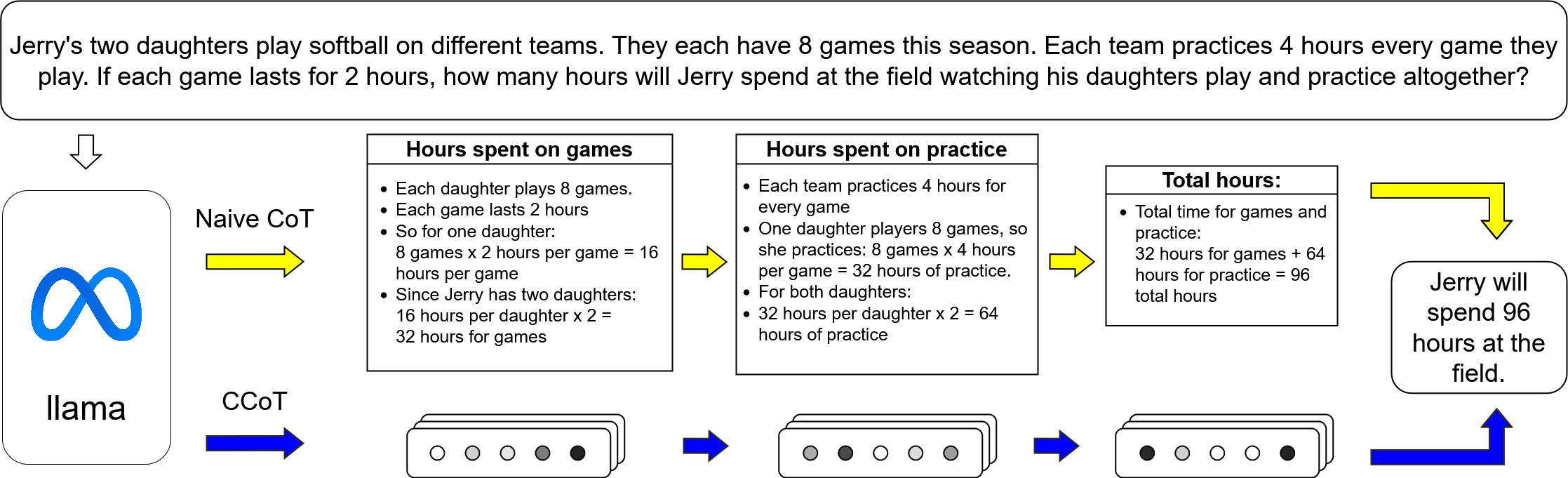}
    \caption{Two approaches to step by step reasoning.
    Chain of Thought (CoT) prompting reasons via discrete language tokens, leading to long sequences that incur significant generation costs. In contrast Compressed Chain of Thought (\textsc{ccot}) elicits reasoning with a short sequence of continuous embeddings, allowing for much greater throughput.}
    \label{fig:intro}
\end{figure*}

Chain-of-Thought (CoT) refers to the Large Language Model (LLM) technique  in which the model simulates the process of thinking out loud by decomposing a complex question into parts and sequentially reasoning through each step. This behavior can be induced by finetuning on a dataset or human feedback \citep{liu2023chainhindsightalignslanguage, puerto2024finetuningdivergentchainsthought}, demonstrating through ICL \citep{wei2023chainofthoughtpromptingelicitsreasoning}, or by providing tuned model instructions \citep{kojima2023largelanguagemodelszeroshot}. While CoT improves the reasoning capabilities of LLMs on a variety of tasks, the improvements come at the cost of a high generation latency. For instance, GPT-4o takes 21.37 seconds to generate a response to the question shown in \cref{fig:intro} with CoT prompting, whereas it can answer the same question without CoT prompting in 2.81 seconds, achieving the same answer with an almost 10x speedup.

Past work has utilized what we term \textit{contemplation tokens} as an alternative to explicit CoT reasoning traces \citep{pfau2024letsthinkdotdot, goyal2024thinkspeaktraininglanguage}. These are additional tokens used to introduce online memory, allowing for additional computations during inference. Instead of generating a reasoning chain entirely of explicit language tokens, the model conditions on a shorter sequence of contemplation tokens (\cref{sec:related}). Contemplation tokens can either be \textit{contentful}, grounded in semantically meaningful text, or \textit{noncontentful}. There are many lines of prior work involving \textit{noncontentful} contemplation tokens drawn from a set of \textit{discrete} tokens; this paper introduces \textit{contentful} contemplation tokens that represent reasoning chains performed in \textit{continuous} space.

Our framework, called Compressed Chain of Thought (CCoT), generates contemplation tokens which are compressed representations of language-based reasoning chains. These contemplation tokens are trained through teacher forcing with respect to the gold hidden states corresponding to full reasoning traces. Our framework can be adapted to pretrained LLMs through LoRA finetuning. Moreover, the variable compression ratio during training allows for need-based adjustments to the performance-efficiency tradeoff by controlling the number of tokens generated during inference. 

The contributions of this paper are as follows:

\begin{enumerate}[nosep]
    \item We finetune pretrained decoder-only LLMs with our new \textsc{ccot} framework and empirically evaluate their performance and throughput on GSM8K;
    \item We establish our framework in context of related work in filler tokens and CoT distillation in terms of performance and efficiency;
    \item We extend theoretical results and demonstrate the computational capacity of \textsc{ccot} contemplations tokens.
\end{enumerate}

\section{Related Work}
\label{sec:related}

\paragraph{Distillation of Knowledge Chains} There has been work in distilling the computations done explicitly when decoding the reasoning chains into computation of the hidden states of the answer \citep{deng2023implicitchainthoughtreasoning, deng2024explicitcotimplicitcot}. Contemporaneous work distills reasoning paths into continuous latent tokens \citep{hao2024traininglargelanguagemodels}. Our method differs in that the contemplation tokens we generate are grounded in text rather than only used as a signal to decode from. This is a critical distinction: our grounding offers the  future potential for decoding the reasoning chain from the compressed representations, allowing for post-hoc human inspection of the LLM's reasoning. Moreover, our method successfully adapts a much larger model (7B compared to 1.5B) using a fraction of data ($\approx 9000$ instances in GSM8K compared to $\approx 400000$ instances in an unreleased augmented GSM8K). 
This suggests that our method can scale better to larger models and is more data efficient.

\paragraph{Filler (Pause) Tokens} Many previous methods have considered decoding \textit{contemplation tokens} to provide an LLM with more compute during inference time. These tokens have gone by many names, such as pause tokens \citep{goyal2024thinkspeaktraininglanguage}, memory tokens \citep{burtsev2021memorytransformer}, filler tokens \citep{pfau2024letsthinkdotdot}, and thinking tokens \citep{herel2024thinkingtokenslanguagemodeling}. These works mainly focus on \textit{noncontentful} contemplation tokens, whose main advantage is their ability to be decoded in parallel, providing the model with a greater computational width without the need to autoregressively decode. 

They have been shown to increase the theoretical computational ability of Transformer LLMs \citep{pfau2024letsthinkdotdot}, but cannot simply be naively applied to induce reasoning gains \citep{lanham2023measuringfaithfulnesschainofthoughtreasoning}. However, through careful pretraining and finetuning, pause tokens have been shown to improve reasoning in both RNNs \citep{herel2024thinkingtokenslanguagemodeling} and 
Transformers \citep{goyal2024thinkspeaktraininglanguage}.  In contrast, the contemplation tokens generated by \textsc{ccot} are contentful as they are compressed representations of reasoning chains. Moreover, they are decoded autoregressively resulting in a greater computational depth as well as width.

\paragraph{Contextual Compression} Transformer LLMs are the de facto standard architecture for modern NLP applications. However, due to the quadratic complexity of its self-attention mechanism, these LLMs are inefficient in tasks with long contexts. Many techniques have been proposed to alleviate this issue, including memory slots \citep{ge2024incontextautoencodercontextcompression}, dynamic compression into nuggets \citep{qin2024dododynamiccontextualcompression}, and low level cache encoding \citep{cachegen}. While most techniques rely on access to the intermediate hidden states of LLMs, there has also been work done in the context of API-only LLMs \citep{jiang2023llmlinguacompressingpromptsaccelerated}. Overall, most of the work in contextual compression deals with efficient compression of known context in order to improve generation latency. The compressed context can then be used in downstream tasks such as retrieval augmented generation or summarization. 

The area of context compression is orthogonal to \textit{comptemplation tokens}. The memory slots of \cite{ge2024incontextautoencodercontextcompression} and the nuggets from \citep{qin2024dododynamiccontextualcompression} encode contentful representations of \textit{known} context, but they are only attended to and never generated during inference. While our work focuses on contentful representations of text, there are two crucial differences: our compressed representations are autoregressively \textit{decoded} during inference and they encode content that is a priori \textit{unknown}. 

\begin{table*}[!t]
    \centering
    \begin{tabular}{A|B|B|C|m{0.34\textwidth} }
        \toprule
        Method & Contentful & Format & Inference & \multicolumn{1}{c}{Additional Notes}\\ \midrule
        Chain of Thought \citep{wei2023chainofthoughtpromptingelicitsreasoning} & Yes & Discrete & Variable-length; \linebreak Autoregressively& Best performing method across reasoning tasks; requires no finetuning; inefficient due to unconstrained sequence length. \\ \midrule
        Filler Tokens \linebreak \citep{pfau2024letsthinkdotdot} & No & Discrete & Fixed-length; \linebreak In parallel & Explicit example of problems only  solvable with contemplation tokens. \\ \midrule
        Pause Tokens \linebreak \citep{goyal2024thinkspeaktraininglanguage} & No & Discrete & Fixed-length; \linebreak In parallel & Best gains seen when contemplation tokens are added during pretraining stage. \\ \midrule
        \textsc{coconut} \linebreak \citep{hao2024traininglargelanguagemodels} & Yes & Continuous & Fixed-length; Autoregressively & Trains contemplation tokens by inserting them after removing reasoning steps. \\ \midrule
        \textsc{ccot} \linebreak (\textbf{Ours}) & Yes & Continuous & Variable-length; Autoregressively & Trains contemplation tokens to approximate compressed reasoning chains. \\ \bottomrule
    \end{tabular}
    \caption{A comparison of different methods to generate \textit{contemplation tokens} in order to introduce extra computation into models during inference. We characterize several aspects of the tokens: (1) contentful, the tokens are either intrinsically contentful or approximate/are distilled from contentful text; (2) format, whether the tokens are drawn from a discrete set of embeddings or are drawn from continuous space; and (3) inference, how the tokens are generated during inference. Any additional notes for each method are included as well.}
\end{table*}

\paragraph{Chain of Thought} Chain-of-thought \cite{wei2023chainofthoughtpromptingelicitsreasoning} was  introduced as a prompting method leveraging in-context learning (ICL) using hand crafted demonstrations. \citet{kojima2023largelanguagemodelszeroshot}  showed  similar behavior could be elicited in a zero-shot context by instructing a model to ``think step-by-step.'' There have been a variety of innovations to CoT, improving on its efficiency and performance. 

In terms of efficiency, novel techniques include getting an LLM to generate steps in parallel from a generated template \citep{ning2024skeletonofthoughtpromptingllmsefficient} and generating reasoning chains in parallel using Jacobi decoding \citep{kou2024cllmsconsistencylargelanguage, zhang2024fastchainofthoughtglancefuture}. In terms of performance, techniques include generating multiple reasoning paths \citep{yao2023treethoughtsdeliberateproblem}, and finetuning on human feedback on generated chains \citep{liu2023chainhindsightalignslanguage, puerto2024finetuningdivergentchainsthought}. Our method differs from prior work in improving the efficiency of CoT as it is not prompt-based and does not rely on Jacobi decoding.

\section{Contemplation Tokens}
\subsection{Preliminaries and Notation}
\label{sec:prelim}
We first give a brief overview of a causal decoder-only language model, equipped with standard Transformer blocks \citep{vaswani2023attentionneed}. Let $V$ be the vocabulary and $w_{1:n}$ be an input sequence, $w_i \in V$. Let $d$ be the hidden dimension, $L$ be the number of layers, and $\theta$ be the parameters of the model. The sequence is first passed through an embedding layer, resulting in a vector $w^0_{1:n}$ where each $w^0_i \in \mathbb{R}^d$. The entire vector $w^0_{1:n} \in \mathbb{R}^{n \times d}$ is then passed through a series of Transformer blocks, $T^i: \mathbb{R}^{n\times d} \to \mathbb{R}^{n\times d}$. We denote the output of each $T^i$ as the \textit{hidden states}. The output of the final Transformer block, $w_{1:n}^{L} \in \mathbb{R}^{n \times d}$, is then passed through the language model head to generate a distribution $p_{1:n}$, $p_i \in \mathbb{R}^{|V|}$, from which the next token is sampled.

\vspace{-1em}

\setlength{\jot}{0.2em}
\begin{align*}
    w_{1:n}^0 &= \textsc{embed}_\theta(w_{1:n}) & \triangleright\textit{embedding layer}\\
    w_{1:n}^{\ell} &= \textsc{attn}_\theta^{\ell-1}(w_{1:n}^{\ell-1}) &\triangleright\textit{ transformer blocks}\\
    p_{1:n} &= \textsc{head}_\theta(w_{1:n}^{\scriptstyle L}) &\triangleright\textit{ pass through lm head}\\
    p(w_{n+1} &\mid w_{1:n}) \sim p_n &\triangleright\textit{ sample next token}
\end{align*}

Notation-wise, any lowercase letter will refer to a \textit{token}, lying in $V$. Any lowercase letter with superscripts will refer to the hidden state after passing through the corresponding layer, lying in $\mathbb{R}^{d}$. Any subscripts refers to a sequence. We will often omit superscripts and instead refer to embeddings with bars ( $\bar{\textrm{ }}$ ) and the entire hidden state with hats ( $\hat{\textrm{ }}$ ). Under this notation, we instead have $\textsc{embed}(w_{1:n}) = \bar{w}_{1:n}$, and with slight abuse of notation, $\textsc{attn}(\bar{w}_{1:n}) = \hat{w}_{1:n}$.

There are also instances where hidden states of an input are computed under two different sets of weights. Suppose we have two sequences of embeddings $\bar{w}$, $\bar{x}$, and we want to compute the hidden states $\hat{w}$ under weights $\theta$ and compute the hidden states of $\hat{x}$ under $\psi$, but crucially conditioned on $\hat{w}$. In this case, we will write $\textsc{attn}_{\theta, \psi}([\bar{w} ; \bar{x}]) = [\hat{w} ; \hat{x}]$ where semicolons indicate vector concatenation.

\subsection{Motivation}
\label{sec:motivation}
In question-answer settings, the input $w_{1:n}$ is a query, and the answer $w_{n+1:n+o} = a_{1:o}$ is generated autoregressively as described above. However as seen in the above description of forward passes through Transformer models, the amount of computations for each query is directly proportional to the query length $n$. As such, we can introduce more computations to the model by attending to an a set of \textit{contemplation tokens}, defined to be any additional tokens generated during inference used to introduce addition memory allowing for additional computations during inference. Rather than solely attending to a query $q = w_{1:n}$, we first can generate a set of contemplation tokens $t = t_{1:m}$ and attend to $[q; t]$ in order to decode a better answer. We emphasize that contemplation tokens are not a novel idea, but a term introduced to unify the many names given to this concept (\cref{sec:related}). 

We define the contemplation tokens $t$ to be \textit{contentful} if either the tokens themselves are semantically contentful or the hidden states corresponding to the contemplation tokens are derived from semantically contentful tokens. We define contemplation tokens that do not fulfill either of these conditions to be \textit{noncontentful}. An example of contentful contemplation tokens are the reasoning chains in chain of thought \citep{wei2023chainofthoughtpromptingelicitsreasoning}; they describe the model's reasoning, fulfilling the first condition of being semantically meaningful. On the other hand, an example of noncontentful contemplation tokens are filler tokens \citep{pfau2024letsthinkdotdot}, as they are simply period characters and their hidden states are trained without any signal from semantically contentful hidden states.

Chain of thought turns out to be the only prior method involving contentful contemplation tokens. The performance gains from utilizing chain of thought are clear; however these benefits are offset by the high generation latency. Suppose the input query consists of $n$ tokens and its corresponding reasoning chain consists of $m$ tokens. As each of the tokens in the reasoning chain need to be autoregressively decoded, the generation of the reasoning chain incurs the cost of $m$ extra passes through the model. Moreover when decoding the answer, the model has to attend to the additional $m$ tokens, resulting in $O(m^2)$ more computations when passing through each attention module. As reasoning chains are often many times longer than the query, the amount of extra computations increases dramatically.\footnote{The average reasoning chain in GSM is 1.5 times longer than their corresponding query. Reasoning chains provided by GPT o1 are hundred of times longer than their query.} 

\subsection{Compressing Reasoning Chains}
\label{sec:compression}

Prior work showed that \textit{noncontentful} contemplation tokens only improved reasoning when the task was computationally bottlenecked \citep{pfau2024letsthinkdotdot} or when the tokens were introduced during pretraining \citep{goyal2024thinkspeaktraininglanguage}. We instead aim to utilize \textit{contentful} contemplation tokens as we believe they would be more applicable to a wider set of tasks. To generate contentful contemplation tokens, we take inspiration from an empirical observation of CoT decoding. 

Suppose we have an input query $w_{1:n}$ and its corresponding reasoning chain $t_{1:m}$. We compute the hidden states of concatenated input as $x = [\hat{w}_{1:n};\hat{t}_{1:m}]$. Decoding an answer conditioned on the hidden states $x$ is equivalent to prompting a language model with the query and chain of thought. Consider taking a subset of $\hat{t}_{1:m}$ along the sequence length axis, denoted as $z_{1:k}$ for some $k << m$. Specifically, for each $1 \leq i \leq k$, there exists some $1 \leq j \leq m$ such that $z_i = \hat{t}_{j}$ at each layer. We observe that training an adapter to decode conditioning on this shortened $[x_{1:n} ;z_{1:k}]$ results in lossless performance on downstream tasks.

Given a query $q$, the naive method utilizing this observation would be to autoregressively generate the reasoning chain $t$, select some learned subset of the encoded hidden states $z$, and train an adapter to decode from the query and subset of hidden states. While this method results in a shorter input sequence when generating the answer and thus reduces the attention computations when decoding the answer, it would still incur the linear cost in generating the reasoning chain. We instead propose learning a module to generate the compressed representations $z$ directly. We denote this module as \textsc{ccot}, short for \textbf{c}ompressed \textbf{c}hain \textbf{o}f \textbf{t}hought, as the contemplation tokens it generates are compressed representations of reasoning chains instead of the full chain.

\section{Approach}
Assume we have a pretrained causal decoder-only language model \textsc{lm}, parameterized by weights $\theta$. We wish to train two modules, \textsc{ccot} and \textsc{decode}, respectively parameterized by weights $\varphi$ and $\psi$. At a high level given a query, $\textsc{ccot}_\varphi$ is responsible for the generation of contemplation tokens. $\textsc{decode}_\psi$ is responsible for decoding the answer conditioned on the initial query and contemplation tokens.

Consider a training instance consisting of a query, full reasoning chain and answer, denoted as $w_{1:n}, t_{1:m}$ and $a_{1:o}$, respectively.  Assume some fixed compression ratio $0 < r < 1$ and let $k = \lceil r \cdot m\rceil$. This compression ratio controls how much the reasoning chains are compressed; $r=1$ corresponds to finetuning on the full reasoning chain while a $r=0$ corresponds to finetuning on just the answer. $\varphi$ and $\psi$ are fine-tuned successively, each initialized from $\theta$.

\subsection{Finetuning $\textsc{ccot}_\varphi$}
\label{sec:phi}
The goal of $\textsc{ccot}_\varphi$ is to generate contemplation tokens. Under \textsc{ccot}, these tokens are a compressed representation of a full reasoning chain, equivalent to a size $k$ subset of the hidden states $\hat{t}_{1:m}$ produced by $\textsc{lm}_\theta$. Since processing all of $t$ and then performing a subset selection still incurs the linear cost of generating all $m$ tokens, $\textsc{ccot}_\varphi$ is thus trained to \textbf{approximate a subset of precomputed hidden states}.

To achieve this, we first precompute the hidden states of the concatenated input. We next use a checkpoint of a scorer used to perform a similar subset selection from \citet{qin2024dododynamiccontextualcompression} in order to perform the subset selection of the hidden states. This scorer is simply a linear layer that takes the embeddings from a predetermined layer $T$ as input, and returns the indices of the selected subset. We discuss other methods of subset selection in \cref{sec:results}.
\begin{align*}
    [\bar{w}_{1:n};\bar{t}_{1:m};\bar{a}_{1:o}] &= \textsc{embed}_\theta([w_{1:n};t_{1:m};a_{1:o}])\\ 
    [\hat{w}_{1:n};\hat{t}_{1:m};\hat{a}_{1:o}] &= \textsc{attn}_\theta([\bar{x}_{1:n};\bar{t}_{1:m};\bar{a}_{1:o}])\\ 
    I &= \textsc{scorer}(\hat{t}_{1:m}^{ \textrm{ }T}) 
\end{align*}
We have that $|I| = k$, and we can index the hidden states $z_{1:k} = \hat{w}_{I}$ to serve as the gold labels. We aim to generate $k$ contemplation tokens $\hat{z}_{1:k}$ conditioned on $w_{1:n}$ under $\varphi$ to approximate the labels, but is not immediately clear what inputs we should use to generate the contemplation tokens. 

A reasonable choice is to use the embeddings of the tokens corresponding to the selected indices, $\hat{w}_I$. This choice would make the hidden state approximation easier due to skip connections in the attention layer: $\hat{w}_I$ are the exact inputs used to compute the hidden states in the noncompressed case. However, the selected tokens are usually punctuation tokens and articles. This choice would require predicting a random sequence of semantically empty tokens when autoregressively decoding as we pass the last layer embeddings $\hat{z}_i^{L}$ through the language model head. Another option would be to learn a single embedding as input to generate each hidden state, but this choice removes the additional computational depth induced by autoregressive decoding.

We instead take inspiration from reasoning over continuous space and use the intermediate hidden layers of the previous contemplation token as input to the next token. Formally, the inputs to generate the contemplation tokens $\hat{z}_{1:k}$ are the embeddings of $z_{0:k-1}^l$ at some fixed layer $l$ where $z_0$ represents the hidden state of the last token of the query. 

\vspace{-2em}
\begin{tikzpicture}[remember picture, overlay]
    \node[anchor=north west] at (0, 0) {
        \begin{minipage}{0.48\textwidth}
        \begin{algorithm}[H]
            \caption{Chain of Thought inference}
            \label{alg:cot}
            \begin{algorithmic}[1]
                \REQUIRE Query $w$, parameters $\theta$
                \STATE $\bar{w} \gets \textsc{embed}_\theta(w)$ \hfill $\triangleright \textit{ embed query}$
                \STATE $\hat{w} \gets \textsc{attn}_\theta(\bar{w})$ \hfill $\triangleright \textit{ compute hidden states}$
                \STATE $z \gets {\color{Yellow}[\langle COT \rangle]}$
                \WHILE{{\color{Yellow}$z_{-1} \neq \langle ANS \rangle$}} 
                    \STATE $[\hat{w};\hat{z}] \gets \textsc{attn}_{{\color{Yellow}\theta}}( [\bar{w}; {\color{Yellow} \textsc{embed}_\theta(z)}])$ 
                    \STATE {\color{Yellow}$x \sim \textsc{head}_\theta (\hat{z}_{-1}^{L})$ } 
                    \STATE $z \gets [z; {\color{Yellow}x}]$ 
                \ENDWHILE
                \STATE $a \gets [\langle ANS \rangle]$
                \WHILE{$a_{-1} \neq \langle EOS\rangle$}
                    \STATE $[\hat{w};\hat{z};\hat{a}] \gets \textsc{attn}_{{\color{Yellow}\theta}} ( [\bar{w}_{1:n} ; {\color{Yellow}\textsc{embed}_\theta([z;a])}])$ 
                    \STATE $x \sim \textsc{head}_{\color{Yellow}{\theta}}(\hat{a}_{-1}^{L})$ \hfill $\triangleright \textit{ sample answer token}$
                    \STATE $a \gets [a ; x]$
                \ENDWHILE
                \STATE \RETURN $a$
            \end{algorithmic}
        \end{algorithm}
        \end{minipage}
    };

    \node[anchor=north west] at (8.5, 0) {
        \begin{minipage}{0.5\textwidth}
        \begin{algorithm}[H]
            \caption{\textsc{ccot} inference}
            \label{alg:ccot}
            \begin{algorithmic}[1]
                \REQUIRE Query $w$, parameters {\color{cyan} $\theta, \varphi, \psi$}, {\color{cyan}autoregressive layer $l$}
                \STATE $\bar{w} \gets \textsc{embed}_\theta(w)$ \hfill $\triangleright \textit{ embed query}$
                \STATE $\hat{w} \gets \textsc{attn}_\theta(\bar{w})$ \hfill $\triangleright \textit{ compute hidden states}$
                \STATE $z \gets {\color{cyan} [\hat{w}_{-1}^l]}$
                \WHILE{ {\color{cyan} $\textsc{end}_\psi(\hat{z}^L)$ is $False$ }} 
                    \STATE $[\hat{w}; \hat{z}] \gets \textsc{attn}_{ {\color{cyan}\theta, \varphi}}( [\bar{w}; {\color{cyan}z}])$ \hfill $\triangleright \textit{ gen. cont. token}$
                    \STATE
                    \STATE $z \gets [z; {\color{cyan} \hat{z}_{-1}^l}]$ \hfill $\triangleright \textit{ append cont. token}$
                \ENDWHILE
                \STATE $a \gets [\langle ANS \rangle]$
                \WHILE{$a_{-1} \neq \langle EOS\rangle$}
                    \STATE $[\hat{w};\hat{z};\hat{a}] \gets \textsc{attn}_{{\color{cyan}\theta, \varphi, \psi}} ( [\bar{w}_{1:n} ; {\color{cyan}z; \textsc{embed}_\theta(a)}])$ 
                    \STATE $x \sim \textsc{head}_{\color{cyan}{\psi}} (\hat{a}_{-1}^{L})$ \hfill $\triangleright \textit{ sample answer token}$
                    \STATE $a \gets [a : x]$
                \ENDWHILE
                \STATE \RETURN $a$
            \end{algorithmic}
        \end{algorithm}
        \end{minipage}

    };
    \node[anchor = north west, text width=\textwidth] at (0, -8) {
        \footnotesize \textit{Figure 2.} Two algorithms for inference with contemplation tokens. \cref{alg:cot} describes the usual chain of thought decoding  while \cref{alg:ccot} describes our method, obtained by replacing the yellow text with the cyan text. While CoT decoding decodes contemplation tokens by passing the LM head across the final hidden state, we use the hidden state at the $l$th layer directly.  
    };
\end{tikzpicture}
\newpage

This choice is quite natural as it generalizes the naive autoregressive decoding strategy (\cref{sec:inf}). We train the parameters of $\varphi$ layer by layer with the following loss:
\begin{equation*}
    \textsc{loss}_\varphi(z_i^{l}, \hat{z}^{l}_i) = \frac1{k}\sum_{i = 1}^k \frac1{\sigma^2(z_i^{l})} \textsc{mse}(z_i^{l}, \hat{z}_i^{l})
\end{equation*}
where $\sigma^2(z)$ denotes the variance of $z$ and \textsc{mse} denotes the usual mean squared error between two vectors. We use a scaled mean squared error in order to normalize hidden states with average $L^1$ norms. These norms differ drastically between different layers within the same model, so the scaled loss allows us to keep a consistent learning rate.

To train the $i$th layer, we pass in the inputs described above and compute forward passes through $i$ Transformer layers, crucially only updating the parameters corresponding to the $i$th layer. When training subsequent layers, the parameters corresponding to the $i$th layer are frozen. This provides a natural segmentation to the approximation task, and we found this improved the generated contemplation tokens.

\subsection{Finetuning $\textsc{decode}_\psi$}
\label{sec:varphi}
We assume a trained module $\textsc{ccot}_\varphi$. Compressed reasoning chains are out of distribution for $\theta$, so we need a separate module in order to effectively condition on the generated contemplation tokens. We train $\textsc{decode}_\psi$ to \textbf{decode the answer from the query and contemplation tokens}.

To do this, we first encode the hidden states of the query and autoregressively generate contemplation tokens $z^*_{1:k}$. Con- \newpage 

trasting the training of $\textsc{ccot}_\varphi$, we perform this generation \textbf{autoregressively} rather than using the precomputed embeddings $z_{0:k-1}^l$ described in \cref{sec:phi} We start by passing in $z_0^l$ and compute the hidden states $\hat{z}_1$. We then autoregressively take $\hat{z}_1^{l}$ as the next input to generate $\hat{z}_2$, until an entire sequence $\hat{z}_{1:k}$ is generated. Then, conditioning on the query and contemplation tokens, we pass in the answer tokens $a_{1:o}$ and compute the next-token distributions $p_{1:o}$.

We finetune $\psi$ with the usual cross-entropy loss given the computed distributions where the probabilities of the next token $a_i$ are drawn from the distribution $p_{i-1}$.
\begin{equation*}
    \textsc{loss}_\psi(a_{1:o}) = -\sum_{i=2}^o \log p(a_i \mid a_{1:i-1}) \sim p_{i-1}
\end{equation*}

\vspace{-1em}

The tokens of $a$ are conditioned on the contemplation tokens $\hat{z}$ generated under $\varphi$. By unfreezing the parameters $\varphi$ when finetuning the parameters $\psi$ using $\textsc{loss}_\psi$, we note that the parameters $\varphi$ receive signal from the downstream task. 

Empirically, we find that this signal is not entirely useful -- downstream performance decreased if all the parameters $\varphi$ are unfrozen. We hypothesize that updating the parameters corresponding to earlier layers affects the autoregressive generation of the contemplation tokens. As such, we find that unfreezing the parameters corresponding to layers \textit{after} the autoregressive layer $l$ ends up improving performance.

$k$ will not be known during test time, so we additionally train a binary classifier $\textsc{end}_\psi$ that takes the final layer of generated hidden states $\hat{z}_i^L$ as input and either predicts whether another contemplation token token should be generated. 
We stop generating contemplation tokens after $h$ tokens. We set $h = 200r$, which would only prematurely terminate less than 3\% of the long tailed distribution of reasoning chains.

\subsection{Inference}
\label{sec:inf}
Assume we have a pretrained causal decoder-only language model parameterized by weights $\theta$. Additionally, assume trained modules $\textsc{ccot}_\varphi$, $\textsc{decode}_\psi$ and the end predictor $\textsc{end}_\psi$. Given a query $w$, we describe inference in \cref{alg:ccot}. We remark that our method to generate contemplation tokens is quite natural; \cref{alg:cot} describes the usual chain of thought inference and the differences are marked, cyan for our method and Yellow for naive CoT decoding.

The most crucial difference is that when \textsc{ccot} generates contemplation tokens (lines 4-7), it uses $l$th layer of the last token's hidden state as a \textit{continuous} next input. In contrast when CoT generates contemplation tokens, it uses the final $L$th layer to do the usual autoregressive decoding described in \cref{sec:prelim}, passing to a \textit{discrete} set of tokens. Moreover, if $m$ is the average length of reasoning chains under $\theta$, \textsc{ccot} will generate on average only $k = \lceil r \times m \rceil $ contemplation tokens, whereas CoT will generate on average all $m$ tokens. 

\subsection{Implementation Details}
We use LORA \citep{hu2021loralowrankadaptationlarge} in order to finetune $\varphi$ and $\psi$ with ranks of 128 and 64, respectively. When generating the gold hidden states, we pass the $T=3$ layer to perform our subset selection and the $l = 15$ as inputs. We also take the hidden state at the $l$th layer to do autoregressive generation of contemplation tokens when finetuning $\psi$ and during inference. We use the decoder-only Transformer architecture of \textsc{llama} for our experiments, taking the \textsc{llama2-7b-chat} checkpoint \citep{touvron2023llama2openfoundation} as our base model. 

\section{Experiments}
\subsection{Experimental Setups}
We evaluate our \textsc{ccot} framework on the reasoning dataset GSM8K \citep{cobbe2021trainingverifierssolvemath}. For the reasoning chains required to train both modules, we use the chains of thought provided with the dataset. We remove all calculator annotations present in the reasoning chain, only keeping the natural language reasoning. We finetune $\varphi$ with precomputed gold states with two compression ratios, $r = [0.05, 0.10]$. We emphasize that the choice of $r$ is a training time decision, $\textsc{ccot}_\varphi$ approximates the hidden states under the fixed compression ratio $r$.

We compare our results to two baselines of $r = [0.0, 1.0]$. These compression ratios are the two extreme values of the compression spectrum we introduce, corresponding to the cases of no reasoning chain and full reasoning chain. We finetune the model with the usual cross entropy loss on the dataset; For $r = 0.0$, the model directly outputs the answer without generating any contemplation tokens. For $r = 1.0$, the model generates the explicit reasoning chain as its contemplation tokens during inference.

Additionally, we compare to \textsc{pause}, a method derived from \citet{goyal2024thinkspeaktraininglanguage}. We finetune the model with no reasoning chains, but for a given ratio $r$, append $k = \lceil r \times m\rceil$ contemplation tokens between the query and answer where $m$ is the length of the reasoning chain. We learn the input embedding of the special token, chosen to be $\langle pause \rangle$. These pause tokens are added to provide the model with an enhanced computational width (See \cref{sec:theory} for further discussion). We evaluate with the same compression ratios $r = [0.05, 0.10]$ to measure the effect of the tokens.
 
\subsection{Results and Discussion}
\label{sec:results}
We provide our main results in \cref{tab:main}. Accuracy refers to the exact match accuracy obtained on the test set with no in-context examples. Decode time refers to the average time to generate an answer to a test set query, measured in seconds by wall clock time on a single Nvidia A100 GPU. 

\begin{table}[!t]
    \centering
    \begin{tabular}{cccc}
        \toprule
        Format & $1/r$ &Acc. (EM) & Decode Time  \\ \midrule
        \textsc{ccot} &$\infty$& 0.089 & 0.33\\
        \rowcolor{LightCyan}\textsc{ccot} &20x & 0.151 & 0.49\\
        \rowcolor{LightCyan}\textsc{ccot} &10x & 0.179 & 0.78\\
        \textsc{ccot} &1x& 0.315 & 8.10 \\ \midrule
        \textsc{pause}& 20x& 0.092& 0.35\\
        \textsc{pause}& 10x& 0.099& 0.37\\ 
        \bottomrule
    \end{tabular}
    \caption{Accuracy and decode time on GSM8K \citep{cobbe2021trainingverifierssolvemath} contrasting our method, \textsc{ccot}, and \textsc{pause} \citep{goyal2024thinkspeaktraininglanguage} each equipped with two different compression ratios against baselines of no compression (full reasoning chains) and infinite compression (no contemplation tokens). Higher accuracy indicates better performance, while lower decode time indicates better efficiency.}
    \label{tab:main}
\end{table}

With a compression ratio of $r = 0.10$, we see a 9 point improvement over the baseline with no contemplation tokens. This accuracy gain is achieved with an only 0.4 second increase in generation time. If we reduce $r$ to 0.05, we still see a sizable 6 point improvement over the baseline, with a generation time increase of only around 0.15 seconds. In contrast, even though the contemplation tokens generated by \textsc{pause} could be decoded faster, they were only able to nominally improve performance. We hypothesize that even though these tokens provide the model with additional computations, reasoning datasets like GSM8K require more sequential computations over parallel ones. Ultimately, our results show equipping a model with dense, contentful contemplation tokens produced by \textsc{ccot} allows the model to reason better than if it had no contemplation tokens, or used a discrete set of noncontentful ones. 

\section{Further Discussion}
\subsection{Hyperparameter Choices}
\paragraph{Varying $r$} As $r$ controls how many contemplation tokens are generated, it makes sense that increasing $r$ would increase both accuracy and decode time. However, we found that accuracy plateaus after a certain threshold, about $r = 0.2$. We hypothesize that this occurs because successive contemplation tokens are autoregressively decoded using the hidden state at the $l$ layer, which propagates an approximation to the next contemplation token generation. We suspect the noise from the approximation errors eventually outweighs the signal provided by the contemplation tokens.

\paragraph{Varying $l$} We find that the choice of $l$ is important -- we were unable to learn good weights for $\varphi$ when $l$ was set close to either 0 or the last layer $L$. We hypothesize that hidden states at earlier layers (small $l$) still incorporate a lot of localized information about the token itself while hidden states at later layers (large $l$) instead incorporate a lot of localized information about the \textit{next} token. As such, we found that $l \approx L / 2$ resulted in the best performance; we hypothesize that the hidden states at intermediate layers encode global information it them suitable for autoregressive decoding scheme we use to generate contemplation tokens.
We provide results with other layer choices in 
\cref{app:hyper}.

\paragraph{Subset selection} We used a learned scorer module to perform the subset selection of the gold hidden states to be emulated by $\varphi$. In practice, we found that simply taking $k$ evenly spaced tokens resulted in a similar performance. However, we note that a module trained to decode from gold hidden states (in the setup described in \cref{sec:compression}) achieves lossless performance compared to decoding from the full reasoning chain, even for small values of $r$. As such, we hypothesize that it is possible to learn a better scorer to identify a subset of hidden states that is easier to emulate; a better approximation of the gold hidden states could lead to lossless performance while only taking a fraction of the time to decode. The observed performance-efficiency tradeoff also likely occurs because it is easier to approximate sequences with less compression.

\subsection{Theoretical Considerations}
\label{sec:theory}

We explore the enhanced computational expressivity offered by contemplation tokens and crucially identify the advantage of decoding contemplation tokens autoregressively rather than in parallel. We provide a few high level intuitions that are formalized in \cref{app:theory}.

\paragraph{Width} Suppose we have a Transformer block $\textsc{attn}$ and an input $\bar{w}_{1:n}$. Computing $\textsc{attn}(\bar{w}_{1:n})$ results in $O(n)$ parallel operations. If we pass in $m$ additional contemplation tokens in parallel and compute $\textsc{attn}(\bar{w}_{1:n+m})$, we now perform $O(n+m)$ parallel operations. The extra computations matter in tasks when the number of parallel operations required is greater than the input sequence length. This can occur when answering succinctly phrased problems that require many parallel operations: ``compute all pairwise sums in the following list'' or ``select all combinations of dependent logical statements that can be mutually true'' Computing pairwise sums of an $n$ element list requires processing $O(n^2)$ parallel computations and computing the validity of all possible logical combinations of $n$ facts requires processing $O(2^n)$ ones. As $n$ grows, introducing contemplation tokens during inference in these \textit{width-bottlenecked} scenarios can allow models to solve additional problems.

\paragraph{Depth} Suppose we have a model consisting of $L$ Transformer blocks, and we generate contemplation tokens autoregressively than in parallel. Passing in $m$ additional contemplation tokens still results in the increased $O(n+m)$ parallel operations, but also resuts in $O(mL)$ sequential operations. These extra computations matter in tasks when the number of sequential operations required is greater than the depth of the model. This can occur in multi-hop question answering tasks or when determining the best move in sequential games such as go and chess. Introducing autoregressively decoded contemplation tokens in these \textit{depth-bottlenecked} scenarios can allow models to solve additional problems.

To collect these observations into a formal theorem, we build from prior work that provides an analysis of the computational power of contemplation tokens decoded in parallel \citep{goyal2024thinkspeaktraininglanguage}. We restate their theorem below:

\begin{theorem}[\textbf{From \citet{goyal2024thinkspeaktraininglanguage}}]
    Assume that the attention module has sufficiently many parameters ($K$) that is much larger than the number of input tokens ($N$). Then there are tasks that $M$ independent computations, where $N < M < K$), such that a 2-layer Transformer can implement the task if and only if it uses contemplation tokens.
\label{thm:1}
\end{theorem}

\vspace{-1em}
Under the same assumptions, autoregressively decoded contemplation tokens can solve a broader class of problems. When the depth of a task $D$ exceeds the number of layers in a model $L$, the model can only represent $L$ steps out of the required $D$ steps. Our intuition is that contemplation tokens can ``save'' the intermediate steps, so autoregressively the model's representation of the $L$th step as the input to the next token allows for the next forward pass to implement another $L$ steps on top of the saved work. Thus, any task of depth $D$ can be solved with an additional $D / L$ contemplation tokens. We note that in \textsc{ccot}, we only pass in the representation of the $l \approx L/2$ step, but this doesn't detract from the asymptotic representational capacity; we simply require an additional $D / l$ tokens instead of $D / L$. We informally state our theorem below, see \cref{thm:3}.

\begin{theorem}
    Assume the conditions in \cref{thm:1}. Then, there are tasks that involve $M$ independent computations of a depth $D>2$ such that a 2-layer Transformer can implement the task if and only if it autoregressively decodes contemplation tokens. 
\end{theorem}

\section{Conclusion} We propose a new framework, \textsc{ccot}, to generate contentful and autoregressively decoded contemplation tokens, a term we introduce to unify the terms given to tokens introducing additional computation to a language model. \textsc{ccot} provides substantial improvements over the baseline as well as methods introduced by prior work. We additionally show how reasoning can be viewed as efficiency-performance tradeoff through the adaptive compression ratio. Overall, our work demonstrates the potential of using contentful contemplation tokens as an alternative to explicit reasoning chains, suggesting a new paradigm of reasoning in continuous space.

\newpage
\bibliography{ccot}
\bibliographystyle{icml2025}

\newpage
\appendix

\section{Varying the autoregressive layer}
\label{app:hyper}

Our method \textsc{ccot} autoregressively generates contemplation tokens by using the hidden state at the $l$th layer at index $i$ as the input embedding at index $i+1$. We show the results of varying this autoregressive layer $l$. We have that $l=0$ corresponds to the embedding layer and $l=L$ corresponds to the final layer prior to passing through the model head.

\begin{table}[h]
    \centering
    \begin{tabular}{>{\centering\arraybackslash}m{0.15\textwidth}c}
        \toprule
        Autoregressive Layer & Accuracy (EM)  \\ \midrule
        \textsc{none} & 0.089 \\
        3 & 0.087 \\
        15 & 0.151 \\
        31 & 0.092 \\
        \bottomrule
    \end{tabular}
    \caption{Accuracy on GSM8K with our method \textsc{ccot} with a compression ratio of $r = 0.05$ when varying the autoregressive layer $l$. \textsc{none} refers to the baseline where no contemplation tokens are decoded during inference.}
\end{table}
\raggedbottom

\section{Further Theoretical Considerations}
\label{app:theory}

In this section, we formalize the two insights outlined in \cref{sec:theory}. We note that an analysis of the enhanced computation width provided by contemplation tokens decoded in parallel was provided by \citet{goyal2024thinkspeaktraininglanguage}. They established a series of assumptions and defined a class of problems involving many parallel operations that a 2-layer Transformer is able to solve only if it leverages contemplation tokens.

We observe that any tasks able to be solved by decoding contemplation tokens in parallel can also be solved by decoding contemplation tokens autoregressively. We thus extend the results from \citet{goyal2024thinkspeaktraininglanguage} by defining a more general set of problems that a 2-layer Transformer is able to solve only if it decodes contemplation tokens autoregressively.

In \textsc{ccot}, contemplation tokens are decoded autoregressively by using the hidden state at the $l$th layer as the next input. As we take $l \approx L/2$, we adapt this framework to a 2-layer Transformer by using the only intermediate layer, $l=1$. We formally introduce the new class of problems and outline the assumptions made by \citet{goyal2024thinkspeaktraininglanguage} below.

\begin{assumption}
    \textit{\textbf{(structure of underlying task)} Assume a vocabulary $\mathcal{V}$ and a embedding dimension of $d$. Let $\circ$ be a genetic 2-ary operator on the embedding space $\mathbb{R}^d$. For a given input length $N$, define the class of functions $\mathcal{F}_{M, K}$ to be the set of all functions $f: \mathcal{V}^N \to \mathcal{V}$ that require applying computing $M$ different $\circ$ operations of depth $K$, followed by a generic aggregation function $g: \mathbb{R}^{M \times d} \to \mathcal{V}$.}
    \label{ass:1}
\end{assumption}
Here, we assume that the vocabulary is passed into the embedding space through an embedding layer prior to the Transformer blocks. This embedding layer is given as $h: \mathcal{V} \to \mathbb{R}^d$. We define $\mathcal{F_{M, K}}$ symbolically as
\begin{align*}
    \mathcal{F}_{M, K} = \Big\{ f: \mathcal{V}^N \to \mathcal{V} \: \Big| \: \exists T_j &\in \mathcal{P}(\{1, \cdots, N\})\\ |T_j| &= K, \forall j \in \{1, \cdots, M\}\\ f(v_1, \cdots, v_N) = g(& \bigcirc_{i \in T_1} \bar{v}_i, \ldots, \bigcirc_{i \in T_M} \bar{v}_i) \Big\}
\end{align*}
Previous work only considered the case of $K=2$ \citep{goyal2024thinkspeaktraininglanguage}, which lends itself well to parallel tasks. This structure extends this case by considering inputs to $g$ that require more sequential computation. Examples of these tasks that require more sequential computations include computing the sums of all triplets in a list of numbers,  multi-hop QA tasks, and generally any problem requiring recursion.

The following assumptions are taken from \citet{goyal2024thinkspeaktraininglanguage}. Further details can be found in the original paper.
\begin{assumption}
    \textit{\textbf{(information bandwidth limit of Transformer hidden states)} We assume that the hidden state corresponding to the $i$th token at any layer can be represented as $(u_i, i)$ by a mapping $h: \mathbb{R}^d \to \mathcal{V} \times \mathbb{N}$.}
    \label{ass:2}
\end{assumption}
\begin{assumption}
    \textit{\textbf{(representational limits of Transformer operations)} Let $v \in \mathbb{R}^{N \times d}$ be a sequence of hidden states. Assume that at every index $i \in \{1, \cdots, N\}, \textsc{attn}(v)_i$ can represent two types of functions.}
    \begin{itemize}
        \item The $\circ$ operation on the hidden states of two arbitrary indices $j, k$. We keep the same assumptions that the self-attention module can select the two indices and the feed-forward module can implement the $\circ$ operation, $\textsc{attn}(v)_i = v_j \circ v_k$.
        \item The aggregating function $g$, the Transformer block can represent $\textsc{attn}(v)_i = (g(v_1, \cdots, v_N), i)$.
    \end{itemize}
    \label{ass:3}
\end{assumption}
\begin{assumption}
    \textit{\textbf{(the capacity of the Transformer block is independent of input length)} We assume that the self-attention module has at least $2T\log T$ parameters for some $T >> N$ and thus can implement any of the $T^T$ possible index mappings This means that the self-attention module can select up to $T$ pairs of .}
    \label{ass:4}
\end{assumption}

\begin{theorem}
    Under the conditions outlined in \cref{ass:1,ass:2,ass:3,ass:4}, the three following statements are true assuming an input sequence of length $N$.
    \begin{itemize}[nosep]
        \item Standard inference with a 2-layer Transformer can only represent the function class $\mathcal{F}_{M, 2}$ for $M \leq N$.
        \item For any $M \leq T$, a 2-layer Transformer that decodes $M-N$ contemplation tokens in parallel can represent the function class $\mathcal{F}_{M, 2}$.  
        \item For any $K > 2$, $MK \leq T$, a 2-layer Transformer that decodes $MK - M$ contemplation tokens autoregressively can represent the function class $\mathcal{F}_{M, K}$.
    \end{itemize} 
    \label{thm:3}
\end{theorem}
\textit{Proof:} To prove the first point, it suffices to show that we can represent the function class $\mathcal{F}_{N, 2}$ under standard inference with an input sequence length of $N$ tokens. We demonstrate this via construction, computing $N$ distinct $\circ$ operations in the first Transformer block, and the aggregation in the second Transformer block. In order to represent all possible choices of pairs, we need to have the $N$ representations of each token at the embedding layer. Expressing $N$ representations must 
use all $N$ indices by \cref{ass:2}. We use the natural choice of using the $i$th index to represent the $i$th token's embedding. By \cref{ass:4}, we can compute the $N$ distinct $\circ$ operations in the first Transformer block, and aggregate them using the second Transformer block. Thus, we show that we can represent $\mathcal{F}_{N, 2}$.

To prove the second point, it suffices to show that we can represent the function class $\mathcal{F}_{N+1, 2}$ by appending a singular contemplation token. We know from \cref{ass:2} that the second Transformer block can aggregate $N+1$ inputs. Assuming $N+1 \leq T$, the first layer can compute the addition $\circ$ operation by \cref{ass:4}. This argument follows by finite induction for any $N+i \leq T$.

For the last point, we observe that the autoregressive inputs will ``save'' an intermediate step which allows it to be conditioned on in the same layer. For instance, to compute $\bar{v}_1 \circ \bar{v}_2 \circ \bar{v}_3$ given the input $v = v_1v_2v_3$, we would let the output of the first block be $\textsc{attn}(\bar{v})_3 = \bar{v}_1 \circ \bar{v}_2$. This gets passed autoregressively as the next input, denoted as $\bar{w}$. Then $\textsc{attn}(\bar{v})_4$ can select the index of the new token and the index of the third token to compute $\bar{w} \circ \bar{v}_3 = \bar{v}_1 \circ \bar{v}_2 \circ \bar{v}_3$.

In order to compute a $\circ$ operation of depth $K$, we need to compute the sequential prefix $\circ$ operations of depth $K-1, \cdots, 2$. This requires a total of $K-1$ extra autoregressively generated contemplation tokens just to compute a single $\circ$ operation of depth $K$. The worst case scenario is that none of the prefix $\circ$ operations are shared, so autoregressively decoding $M(K-1)$ contemplations will allow us to compute all $M$ $\circ$ operations. We have at most $MK$ total tokens, so given $MK \leq T$, we can compute the desired $\circ$ operations by \cref{ass:4}. $\square$

\end{document}